\documentclass[10pt,twocolumn,letterpaper]{article}

\usepackage[pagenumbers]{cvpr} 

\usepackage{graphicx}
\usepackage{amsmath}
\usepackage{amssymb}
\usepackage{booktabs}
\usepackage{enumitem}
\usepackage{bbding}
\usepackage{caption}
\usepackage{multirow}
\usepackage{makecell}
\usepackage{floatrow}
\usepackage{flushend}
\newfloatcommand{capbtabbox}{table}[][\FBwidth]

\usepackage[pagebackref,breaklinks,colorlinks]{hyperref}

\usepackage[capitalize]{cleveref}
\crefname{section}{Sec.}{Secs.}
\Crefname{section}{Section}{Sections}
\Crefname{table}{Table}{Tables}
\crefname{table}{Tab.}{Tabs.}

\begin{document}

%%%%%%%%% TITLE - PLEASE UPDATE
\title{Sparse4D: Multi-view 3D Object Detection with Sparse Spatial-Temporal Fusion}

\author{Xuewu Lin, Tianwei Lin, Zixiang Pei, Lichao Huang, Zhizhong Su\\
Horizon Robotics\\
{\tt\small xuewu.lin@horizon.ai}
}
\maketitle

%%%%%%%%% ABSTRACT
\begin{abstract}

Bird-eye-view (BEV) based methods have made great progress recently in multi-view 3D detection task.
Comparing with BEV based methods, sparse based methods lag behind in performance, but still have lots of non-negligible merits.
To push sparse 3D detection further,
in this work, we introduce a novel method, named Sparse4D, which does the iterative refinement of anchor boxes via sparsely sampling and fusing spatial-temporal features.
(1) \textbf{Sparse 4D Sampling}: for each 3D anchor, we assign multiple 4D keypoints, which are then projected to multi-view/scale/timestamp image features to sample corresponding features;
(2) \textbf{Hierarchy Feature Fusion}: we hierarchically fuse sampled features of different view/scale, different timestamp and different keypoints to generate high-quality instance feature.
In this way, Sparse4D can efficiently and effectively achieve 3D detection without relying on dense view transformation nor global attention, and is more friendly to edge devices deployment.
Furthermore, we introduce an instance-level depth reweight module to alleviate the ill-posed issue in 3D-to-2D projection.
In experiment, our method outperforms all sparse based methods and most BEV based methods on detection task in  the nuScenes dataset. Code is available at \url{https://github.com/linxuewu/Sparse4D}.
\end{abstract}

\begin{figure}
  \centering
  \includegraphics[width=0.99\linewidth]{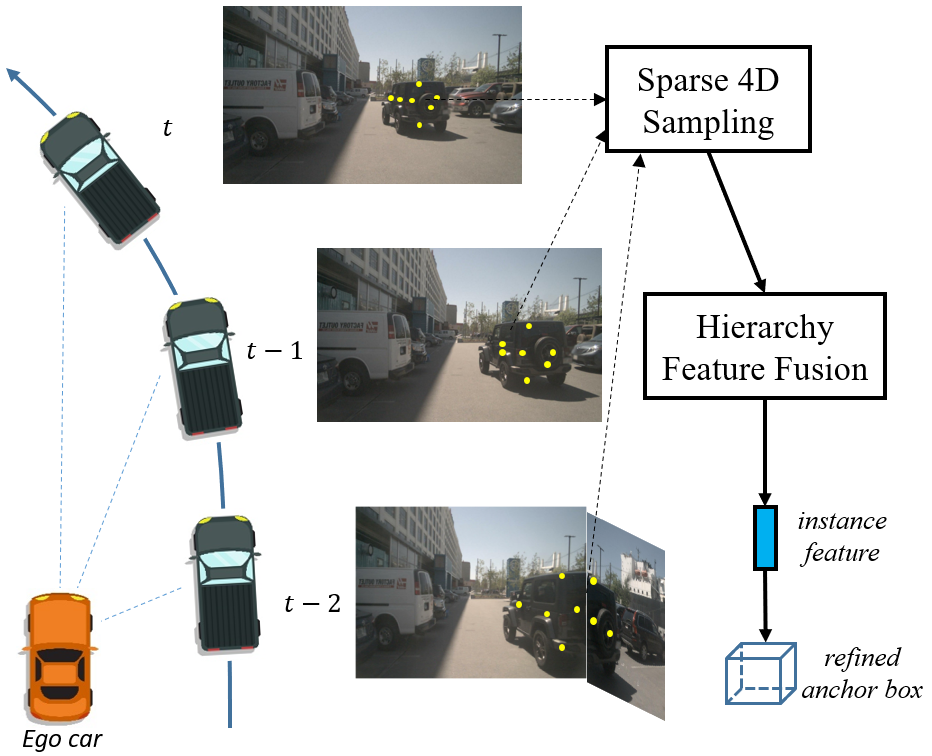}
  \caption{Overview of the Sparse4D. For each candidate anchor instance, we sparsely sampling multi-timestamp/view/scale features of multiple keypoints, then hierarchically fuse these feature as instance feature for precise anchor refinement.}
  \label{fig:overview}
\end{figure}

%%%%%%%%% BODY TEXT
\section{Introduction}
\label{sec:intro}

Multi-view visual 3D perception plays a critical role in autonomous driving systems, especially for low-cost deployment.
Compared with Lidar modality, cameras can provide valuable visual cues for long-range distance detection and vision-only element identification.
However, without explicit depth cues, 3D perception from 2D images is an ill-posed issue, leading to a long-standing challenge of how to properly fuse multi-camera images to address  3D perception tasks such as 3D detection.
There are two mainstream categories of recent methods: the BEV-based methods and the sparse-based methods.

BEV-based methods~\cite{bevformer,zhang2022beverse,li2022bevdepth,huang2021bevdet,liang2022bevfusion,solofusion} address 3D detection via converting multi-view image features into an unified BEV space, and achieve excellent performance promotion.
However, besides the advantages of the BEV fashion, there still exist some unavoidable disadvantages  as follows:
(1) The image-to-BEV perspective transformation requires dense feature sampling or rearrangement,
which is complex and computationally expensive for low-cost edge devices deployment;
(2) The maximum perception range is limited by the size of BEV feature map, making it difficult to trade off among perception range, efficiency and precision;
(3) The height dimension is compressed in BEV feature with losing of texture cues. Thus BEV features are  incompetent for some perception tasks such as sign detection.

Different from BEV based methods, the sparse based algorithms~\cite{wang2022detr3d,chen2022graph,shi2022srcn3d} do not require a dense perspective transformation module, but directly sample sparse feature for 3D anchors refinement, thus can alleviate above issues.
Among them, the most representative sparse 3D detection method is DETR3D~\cite{wang2022detr3d}. However, its model capacity is limited since DETR3D only sample feature of a single 3D reference point for each anchor query.
Recently, SRCN3D~\cite{shi2022srcn3d} utilizes RoI-Align~\cite{he2017mask} to sample multi-view feature, but is not efficient enough and cannot precisely align feature points from different views.
Meanwhile, existing sparse 3D detection methods have not taken advantage of rich temporal context, and  have a significant performance gap compared with state-of-the-art BEV based methods.

In this work, we devote our best effect to expand the limit of sparse based 3D detection.
To address existing issues,
we introduce a novel framework named Sparse4D, which utilizes multiple keypoints distributed in the region of 3D anchor box to sample feature.
Compared with the single point manner \cite{wang2022detr3d} and the RoI-Align manner \cite{shi2022srcn3d}, our sampling manner has two main advantages:
(1) can efficiently extract rich and complete context inside each anchor box;
(2) can be simply extend to temporal dimension as 4D keypoints, then can effectively align temporal information.
With 4D keypoints, as illustrated in ~\cref{fig:overview}, Sparse4D first performs multi-timestamp, multi-view and multi-scale for each keypoint.
These sampled features then go through a hierarchical fusion module to generate high-quality  instance feature for 3D box refinement.
Further, to alleviate the ill-posed issue of camera-based 3D detection and improve the perceptual performance, we explicitly add an instance-level depth reweight module,
where the instance feature is reweighted by depth confidence sampled from predicted depth distribution.
This module is trained in a sparse way without additional Lidar point cloud surpervision.

In summary, our work have four main contributions:
\begin{itemize}[leftmargin=*,itemsep=0pt,partopsep=1pt,parsep=\parskip, topsep=5pt]
\item To the best of our knowledge, our proposed Sparse4D is the first sparse multi-view 3D detection algorithm with temporal context fusion, which can efficiently and effectively align spatial and temporal visual cues to achieve precise 3D detection.
\item We propose a deformable 4D aggregation module that can flexibly complete the sampling and fusion of multi-dimensional (point, timestamp, view and scale) features.
\item We introduce a depth reweight module to  alleviate the ill-posed issue in image-based 3D perception system.
\item On the challenging benchmark - nuScenes dataset, Sparse4D outperforms all existing sparse based algorithms and most BEV-based algorithms on 3D detection task, and also performs well on tracking task.
\end{itemize}
%------------------------------------------------------------------------
\section{Related Work}
\label{sec:related}

\begin{figure*}
  \centering
  \includegraphics[width=0.95\textwidth]{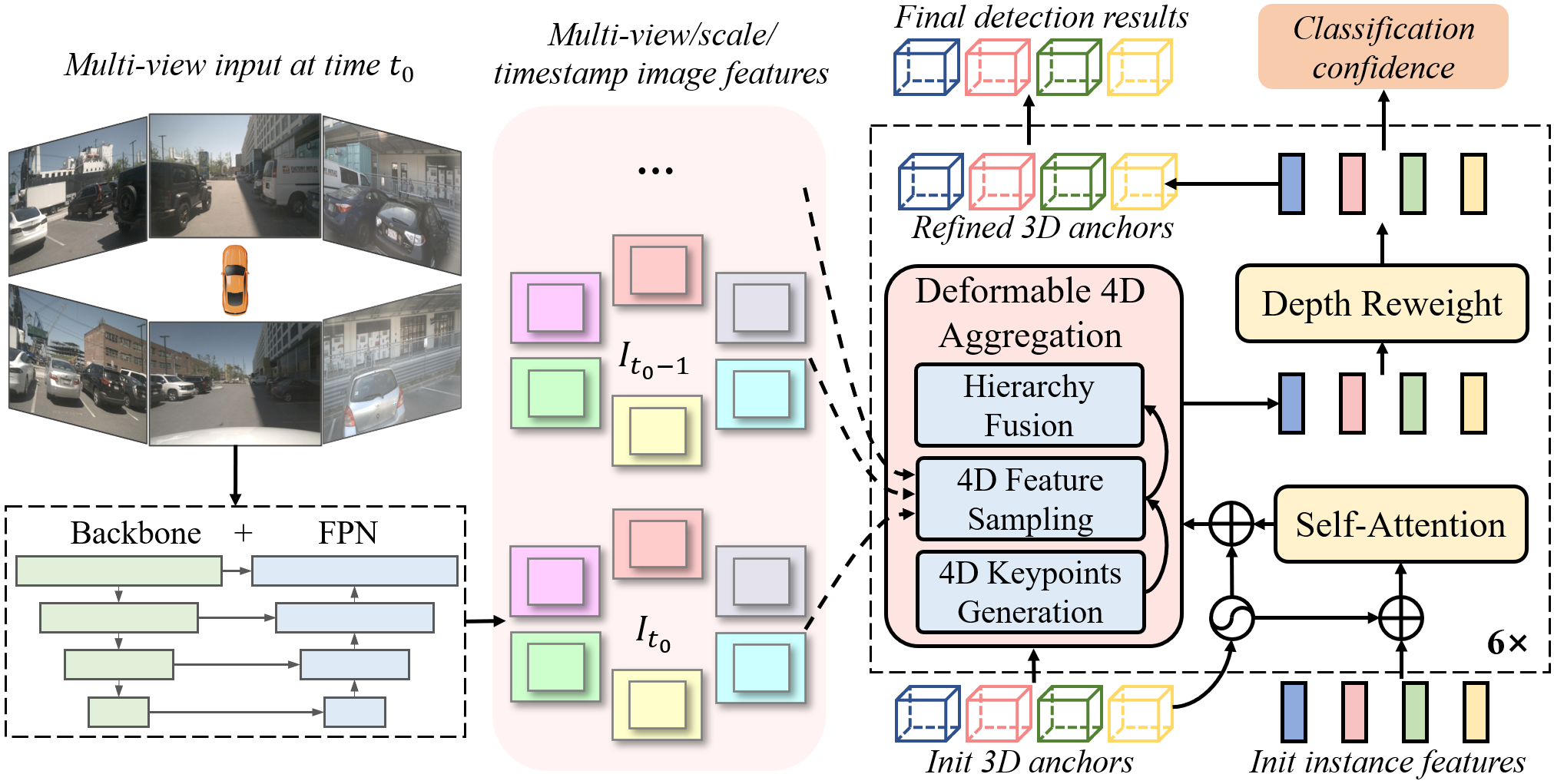}
  \vspace{-0.1cm}
  \caption{Overall architecture of Sparse4D. Taking multi-view images as input, we first extract multi-timestamp/view/scale feature maps with the image feature encoder. The decoder contains multiple refinement modules with independent parameters, which take image feature maps, instance features and 3D anchors as input, and continuously refines the 3D anchors to obtain accurate detection results.}
  \vspace{-0.1cm}
  \label{fig:overall_architecture}

\end{figure*}
%-------------------------------------------------------------------------

\subsection{Sparse Object Detection}
Early object detection methods~\cite{ren2015faster,liu2016ssd,tian2019fcos,duan2019centernet,tan2020efficientdet} used dense predictions as output, and then utilized non-maxima suppression (NMS) to process those dense predictions.
DETR~\cite{DETR} introduces a new detection  paradigm that utilizes set-based loss and transformer to directly predict sparse detection results.
DETR performs cross attention between object-query and global image context, leading to heavy computation cost and difficulty in convergence.
Due to the use of global cross attention, DETR cannot be regarded as a pure sparse method.
Deformable DETR~\cite{deformableDETR} then modifies DETR and proposes a local cross attention based on reference points, which accelerates the model convergence and reduced computational complexity.
Sparse R-CNN~\cite{sparsercnn} proposes another sparse detection framework based on the idea of region proposal. The network structure is extremely simple and effective, showing the feasibility and superiority of sparse detection.
As the extension of 2D detection, many 3D detection methods have recently paid more  attention to these sparse paradigms, such as MoNoDETR~\cite{zhang2022monodetr}, DETR3D~\cite{wang2022detr3d}, Sparse R-CNN3D~\cite{shi2022srcn3d}, SimMOD~\cite{zhang2022simple}, etc.
%-------------------------------------------------------------------------

\subsection{Monocular 3D Object Detection}
The monocular 3D detection algorithm takes a single image as input and outputs the 3D bounding box of the objects. Since the image does not contain depth information, this problem is ill-posed, and is more challenging compared with 2D detection.
%thus has a significant increase in difficulty relative to 2D detection.
%
FCOS3D~\cite{wang2021fcos3d} and SMOKE~\cite{liu2020smoke} is extended based on a single-stage 2D detection network, using a fully convolution network to directly regress the depth of each object.
~\cite{wang2019pseudo,weng2019monocularpseudo,qian2020end2endpseudo} convert the 2D image into the 3D pseudo point cloud signal with monocular depth estimation results, and then use the LiDAR-based detection network to complete the 3D detection.
    OFT~\cite{OFT} and CaDDN~\cite{reading2021categorical} transform the dense 2D image feature into BEV space with the help of the view transformation module and then send the BEV feature to the detector to complete 3D object detection.
    The difference is that OFT uses the 3D to 2D inverse projection relationship to complete the feature space transformation, while CaDDN is based on the 2D to 3D projection, which is more like a pseudo-LiDAR method.
%-------------------------------------------------------------------------
\subsection{Multi-view 3D Object Detection}
    Dense algorithms are the main research direction of multi-view 3D detection, which use dense feature vectors for view transformation, feature fusion or box prediction.
    Currently, BEV-based methods are the main part of dense algorithms.
    BEVFormer~\cite{bevformer} adopts deformable attention to complete the BEV feature generation and dense spatial-temporal feature fusion.
    BEVDet~\cite{huang2021bevdet,huang2022bevdet4d} uses lift-splat operation~\cite{philion2020lift} to achieve the view transformation.
    On the basis of BEVDet, BEVDepth~\cite{li2022bevdepth} adds explicit depth supervision, which significantly improves the accuracy of the detection.
    BEVStereo~\cite{li2022bevstereo} and SOLOFusion~\cite{solofusion} introduce temporal stereo technology into 3D detection, further improving the depth estimation effect.
    PETR~\cite{liu2022petr,liu2022petrv2} utilizes 3D position encoding and global cross attention for feature fusion, but the global cross attention is computationally expensive.
    Like vanilla DETR~\cite{DETR}, PETR cannot be regarded as a purely sparse method.
    DETR3D~\cite{wang2022detr3d} is a representative work of sparse methods, which performs feature sampling and fusion based on sparse reference points.
    Graph DETR3D~\cite{chen2022graph} follows DETR3D and introduces a graph network to achieve better spatial feature fusion, especially for multi-view overlapping regions.

%------------------------------------------------------------------------
\section{Methodology}
\subsection{Overall Framework}
As shown in \cref{fig:overall_architecture}, Sparse4D  conforms to an encoder-decoder structure.
The image encoder is used to extract image features with shared weights, which contains a backbone (\eg, ResNet~\cite{resnet} and VoVNet~\cite{vovnet}) and a neck (\eg, FPN~\cite{fpn}).
Given $N$ view input images at time $t$, the image encoder extracts multi-view multi-scale feature maps as $I_t=\left\{I_{t,n,s} | 1 \le s \le S, 1 \le n \le N \right\}$.
To exploit temporal context, we extract image feature of recent $T$ frames as image feature queue $ I = \left\{I_t\right\}_{t=t_s}^{t_0}$, where $t_s = t_0 - (T -1)$.

Then, the decoder predicts detection results in an iteratively refinement fashion, which contains a series of  refinement modules and a classification head for predicting final classification confidences in the end.
Each refinement module takes image feature queue $I$, 3D anchor boxes $B \in \mathbb{R}^{M \times 11}$ and corresponding instance features $F \in \mathbb{R}^{M \times C}$ as inputs, then outputs refined 3D boxes with updated instance features.
Here, $M$ is the number of anchors and $C$ is the feature channel number. The format of an anchor is
\[ \left\{ x, y, z, \ln w, \ln h, \ln l, \sin{yaw}, \cos{yaw}, vx, vy, vz\right\}, \]
All 3D anchors are set in a unified 3D coordinate system(\eg central LiDAR coordinate).

\begin{figure*}
  \centering
  \includegraphics[width=0.99\textwidth]{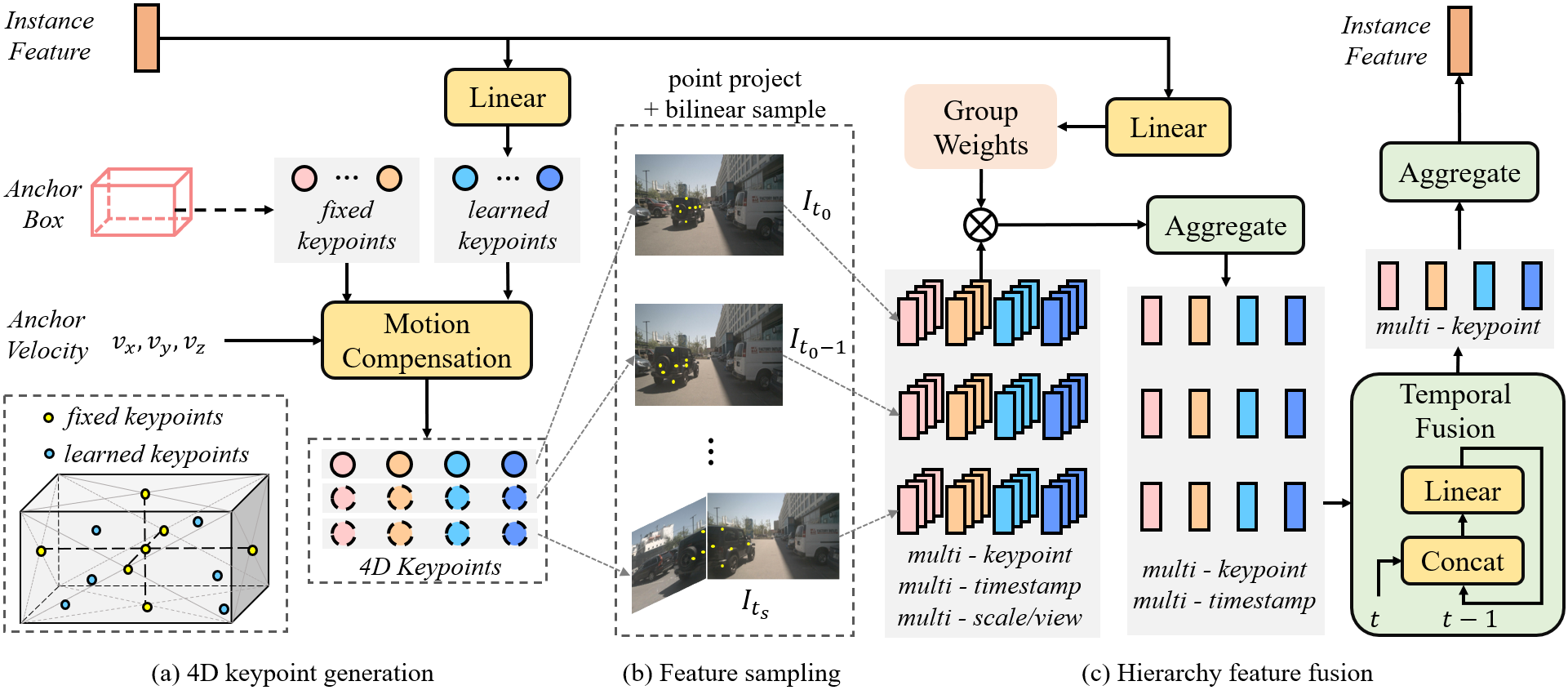}
  \caption{Detailed flowchart of the Deformable 4D Aggregation module. In this module, we extract high-quality instance feature in three steps: (a) for each anchor, generate multiple 4D keypoints; (b) project 4D keypoints to multi-timestamp/view/scale image feature maps and sample corresponding features and (c) hierarchically fuse keypoint features with predicted weights to generate fused instance feature.
  }
  \label{fig:refinement_module}
\end{figure*}

In each refinement module, we first adopt self-attention to realize the interaction between instances, with the embedding of anchor parameters added before and after.
Then, we conduct deformable 4D aggregation (\cref{deformable_agg}) to fuse multi-view, multi-scale, multi-timestamp and multi-keypoint features.
Furthermore, we introduce a depth reweight module (\cref{depth_reweight}) to alleviate the ill-posedness issue in image-based 3D detection.
Finally, a regression head is used to refine the current anchor via predicting the offset between ground truth and the current anchor.

%-------------------------------------------------------------------------
\subsection{Deformable 4D Aggregation}
\label{deformable_agg}

The quality of instance features have a critical impact on the overall sparse perception system.
To address this, as demonstrated on \cref{fig:refinement_module}, we introduce the deformable 4D aggregation module to obtain high-quality instance features with sparse feature sampling and hierarchy feature fusion.

%\noindent
\textbf{4D Keypoints Generation.}
For the $m$-th anchor instance, we assign $K$ 4D keypoints as $P_m \in \mathbb{R}^{K \times T \times 3}$, which are composed of $K_F$ fixed keypoints and $K_L$ learnable keypoints.
As shown in \cref{fig:refinement_module}(a), at current timestamp $t_0$, we first put fixed keypoints $P^F_{m, t_0}$ directly on the stereo center and the six faces center of the anchor box.
%$K = K_F + K_L$
Then, unlike fixed keypoints, the learnable keypoints vary with different instance features, which allow the neural network to find the most representative feature of each instance.
Given instance feature $F_m$ with anchor box embedding added, the learnable keypoints $P^L_{m, t_0}$ are generated by the following formula through a sub-network $\Phi$:

\begin{equation}
      D_{m} = \textbf{R}_{yaw} \cdot \left[\textbf{sigmoid}\left(\Phi(F_m) \right) - 0.5\right] \in \mathbb{R}^{K_L \times 3}
      \label{eq:fixkeypoints_1}
 \end{equation}
\begin{equation}
      P^L_{m, t_0} =D_{m} \times \left[w_m, h_m, l_m\right] + \left[x_m, y_m, z_m\right]
      \label{eq:fixkeypoints_2}
 \end{equation}
where $\textbf{R}_{yaw}$ denotes the rotation matrix of $yaw$.

Temporal features are crucial for 3D detection and can improve depth estimation accuracy. Therefore, after getting the 3D keypoints of the current frame, we extend them to 4D to prepare for temporal fusion.
For a past timestamp $t$, we first build a constant velocity model to shift each 3D keypoints in the 3D coordinate system of the current frame.
\begin{equation}
%P_{t-h}' = P_{t} - \frac{{\rm d}t_h}{{\rm d}t_1} \left[{\rm d}x, {\rm d}y, {\rm d}z\right], 1 < h \le H
P_{m,t}' = P_{m, t_0} - d_t \cdot (t_0 - t) \cdot \left[vx_m, vy_m, vz_m\right]
      \label{eq:constantvel}
 \end{equation}
where $d_t$ is the time interval between two adjacent frames.
Then, we use the ego vehicle motion information to convert $P_{m, t}'$ to the coordinate system of the past $t$ frame.
 \begin{equation}
P_{m, t} = \textbf{R}_{t_0 \rightarrow t} P_{m, t}' + \textbf{T}_{t_0 \rightarrow t}
    \label{eq:temporalproject}
\end{equation}
where $\textbf{R}_{t_0 \rightarrow t}$ and $\textbf{T}_{t_0 \rightarrow t}$ represent the rotation matrix and translation of the ego vehicle from current frame $t_0$ to frame $t$, respectively.
In this way, we can finally construct 4D keypoints as $P_m = \left\{P_{m, t} \right\}_{t=t_s}^{t_0}$.

%\noindent
\textbf{Sparse Sampling.}
Based on the above 4D keypoints $P$ and the image feature maps queue $F$, sparse features with strong representation ability can be efficiently sampled.
First, the 4D keypoints are projected onto the feature maps through the transformation matrix $\textbf{T}^{\rm cam}$.
\begin{equation}
    P_{t,n}^{\rm img} = \textbf{T}_{n}^{\rm cam} P_{t}, 1 \le n \le N
    \label{eq:project2d}
\end{equation}

Then, we conduct multi-scale feature sampling for each view and each timestamp via bilinear interpolation:
\begin{equation}
  f_{m,k,t,n,s} = \textbf{Bilinear}\left(I_{t,n,s}, P_{m,k,t,n}^{\rm img}\right)
  \label{eq:featuresampling}
\end{equation}
where the subscript $m$, $k$, $t$, $n$ and $s$ here indicate the indices of anchor, keypoint, timestamp, camera and feature map scale, respectively. So far, we have obtained multi-keypoints, timestamp, view and scale feature vectors $f_m \in \mathbb{R}^{K \times T \times N \times S \times C}$  for the $m$-th candidate detection anchor, where $C$ is the number of feature channels.

%\noindent
\textbf{Hierarchy Fusion.}
To generate high quality instance feature, we fuse the above features vectors $f_m$ in a hierarchical manner.
As shown in \cref{fig:refinement_module}(c), for each keypoint, we first aggregate features in different view and scale with predicted weights and then conduct temporal fusion with sequence linear layers. Finally, for each anchor instance,  we fuse multi-point features to generate instance feature.

Specifically, given instance feature $F_m$ with anchor box embedding added, we first predict group weighting coefficients through a linear layer $\Psi$ as:
\begin{equation}
  W_{m} = \Psi \left( F_m \right) \in \mathbb{R}^{K \times N \times S \times G}
  \label{eq:weights}
\end{equation}
where $G$ is the number of groups to divide features by channels. With this, we can aggregate channels of different groups with different weights, which is similar to group  convolution~\cite{krizhevsky2017imagenet}.
%and the head in the attention mechanism~\cite{vaswani2017attention}.
We sum the weighted feature vectors  for each group along the scale and view dimensions, and then concatenate the groups to obtain the new features $f^{'}_{m,k,t}$.
\begin{equation}
  f_{m,k,t,i}^{'} = \sum_{n=1}^{N}\sum_{s=1}^{S}{W_{m,k,n,s,i} f_{m,k,t,n,s,i}}
%f_{m,k,t-h,i}^{'} = \sum_{n=1}^{N}\sum_{s=1}^{S}{W_{m,k,n,s,i} f_{m,k,t-h,n,s,i}}
  \label{eq:fusion-view-camera}
\end{equation}
\begin{equation}
  f_{m,k,t}^{'} = \left[f_{m,k,t,1}^{'}, f_{m,k,t,2}^{'} , ... , f_{m,k,t,G}^{'}\right]
  \label{eq:concatgroup}
\end{equation}
The subscript $i$ above is the index of the group, and $\left[,\right]$ represents the concatenate operation. Next, with concatenation operations and a linear layer $\Psi_{temp}$, the timestamp dimension of the  features $f^{'}_{m,k,t}$ will be fused in a sequential manner.
\begin{equation}
\begin{aligned}
& f_{m, k, t_s}^{''} = f_{m, k, t_s}^{'} \\
& f_{m, k, t}^{''} =  \Psi_{temp} \left(\left[f_{m, k, t}^{'}, f_{m, k, t-1}^{''}\right]\right) \\
& f_{m, k}^{''} =f_{m, k,t_0}^{''}=  \Psi_{temp} \left(\left[f_{m, k, t_0}^{'}, f_{m, k, t_0-1}^{''}\right]\right)
\end{aligned}
\end{equation}

The multi-keypoint features $ f_{m, k}^{''} $ after temporal fusion will be summed to complete the final feature aggregation and get the updated instance feature as:
\begin{equation}
  F_{m}^{'} =\sum_{k=1}^{K} f_{m,k}^{''}
  \label{eq:fusion-multi-keypoint}
\end{equation}
%-------------------------------------------------------------------------

\begin{figure}[t]
\centering
\includegraphics[width=1.0\linewidth]{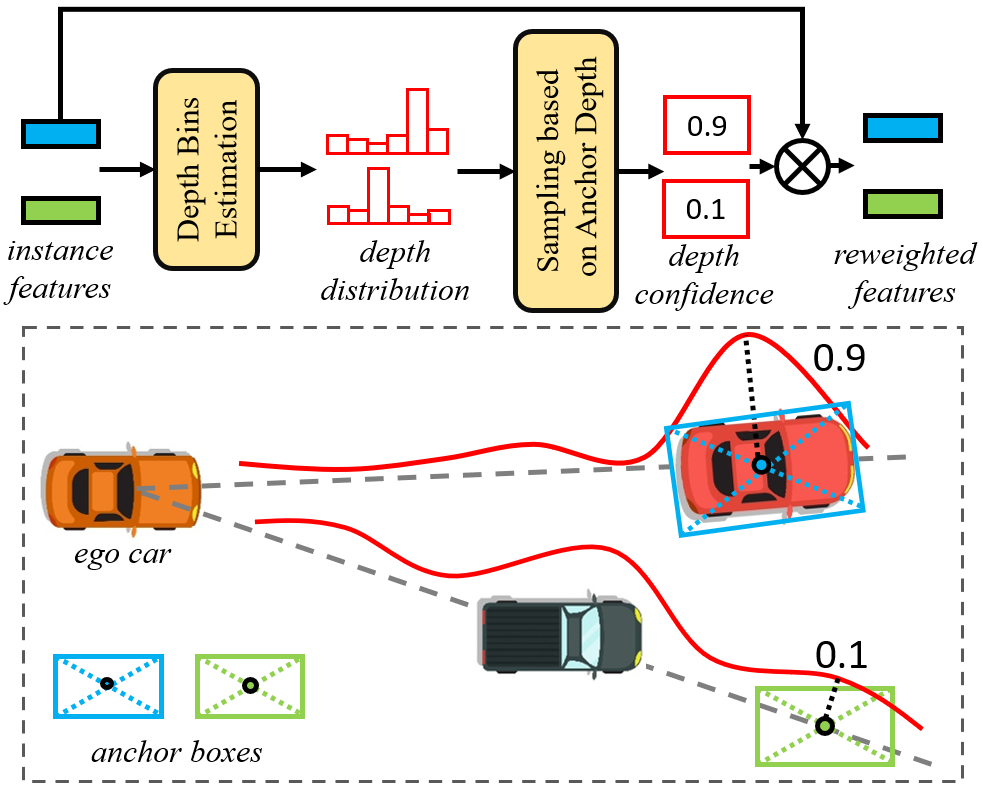}
\caption{Illustration of depth reweight module. For each anchor instance, we estimate the depth distribution along the ray from the camera to the center of anchor box, then sample the depth confidence in the depth of anchor box. The sampled depth confidence is used to reweight instance feature.
}
\label{fig:depthnet}
\end{figure}

\subsection{Depth Reweight Module}
\label{depth_reweight}
This 3D to 2D transformation (\cref{eq:project2d}) has a certain ambiguity, that is, different 3D points may correspond to the same 2D coordinates. For different 3D anchors, the same features may be sampled (see \cref{fig:depthnet}), which increases the difficulty of neural network fitting. To alleviate this problem, we incorporate an explicit depth estimation module $\Psi_{depth}$, which consists of multiple MLPs with residual connections. For each aggregated feature $F_{m}'$, we estimate a discrete depth distribution, and use the depth of center point of 3d anchor box to sample the corresponding confidence $C_{m}$, which will be used to reweight the instance feature.
\begin{equation}
  C_{m} = \textbf{Bilinear}\left(\Psi_{depth}(F_{m}^{'}), \sqrt{x_{m}^2+y_{m}^2}\right)
  \label{eq:depthsample}
\end{equation}
\begin{equation}
  F_{m}^{''} = C_{m} \cdot F_{m}^{'}
  \label{eq:depthmul}
\end{equation}

In this way, for those instances whose 3D center points are far from the ground truth in the depth direction, even if the 2D image coordinates are very close to the ground truth, the corresponding depth confidence tends to zero. Thus the corresponding instance feature $F_m''$ is punished after reweighting also tend to $\textbf{0}$.
Incorporating an explicit depth estimation module can help the visual perception system to further improve the perception accuracy.
Also, the depth estimation module can be designed and optimized as a separate part to facilitate model performance.
%-------------------------------------------------------------------------
\begin{table*}%{\textwidth}
\begin{floatrow}
\capbtabbox{
  \begin{tabular}{@{}cc|ccc@{}}
    \toprule
    DRM & LKP & mAP$\uparrow$ & mAOE$\downarrow$ & NDS$\uparrow$ \\
    \midrule
    \XSolidBrush & \XSolidBrush & 0.432 & 0.408 & 0.533 \\
    \checkmark & \XSolidBrush & 0.431 & 0.381 & 0.537\\
    \XSolidBrush & \checkmark & 0.432 & 0.379 & 0.537 \\
    \checkmark & \checkmark & \textbf{0.436} & \textbf{0.363} & \textbf{0.541} \\
    \bottomrule
  \end{tabular}}{
  \caption{Ablation study of the influence of Depth Reweight Module (DRM) and Learnable Key Points (LKP) on detection effect.}
  \label{tab:ablation-DRM-LKP}}

\capbtabbox{
  \begin{tabular}{@{}ccc|cccc@{}}
    \toprule
    $H$ & Ego & Object & mAP$\uparrow$ & mATE$\downarrow$ & mAVE$\downarrow$ & NDS$\uparrow$ \\
    \midrule
    0 & \textbf{-} & \textbf{-} & 0.322 & 0.747 & 0.890 & 0.401 \\
    3 & \XSolidBrush & \XSolidBrush & 0.334 & 0.759 & 0.682 & 0.424 \\
    3 & \checkmark & \XSolidBrush & \textbf{0.376} & \textbf{0.678} & 0.398 & 0.488 \\
    3 & \checkmark & \checkmark & 0.373 & 0.687 & \textbf{0.329} & \textbf{0.495} \\
    \bottomrule
  \end{tabular}}{
  \caption{Ablation study of the influence of Motion Compensation setting. In this experiment, the input image size is set to $320\times 800$ and learnable keypoints are removed. $H$ is number of history frames.}
  \label{tab:ablationmotion}}
\end{floatrow}
\end{table*}

\subsection{Training}
We sample video clips with $T$ frames to train the detector end to end. The time interval between consecutive frames is randomly sampled in $\left\{d_t, 2d_t\right\}$  ($d_t\approx 0.5$). Following DETR3D~\cite{wang2022detr3d}, the Hungarian algorithm is used to match each ground truth with one predicted value.
The loss includes three parts: classification loss, bounding box regression loss and depth estimation loss:
    \begin{equation}
      L = \lambda_{1} L_{cls} + \lambda_{2} L_{box} + \lambda_{3} L_{depth}
      \label{eq:loss}
    \end{equation}
 where $\lambda_1$, $\lambda_2$ and $\lambda_3$ are weight terms to balance the gradient.
 We adopt focal loss~\cite{lin2017focal} for classification, $L_1$ loss for bounding box regression, and binary cross entropy loss for depth estimation.
 In depth reweight module, we directly use the depth of  the labeled bounding box center as the ground truth to supervise per-instance depth.
Since we only estimate per-instance depth rather than dense depth, the training process gets rid of the dependence on LiDAR data.

%------------------------------------------------------------------------
\section{Experiment}
\begin{figure*}
  \centering
    \includegraphics[width=0.99\textwidth]{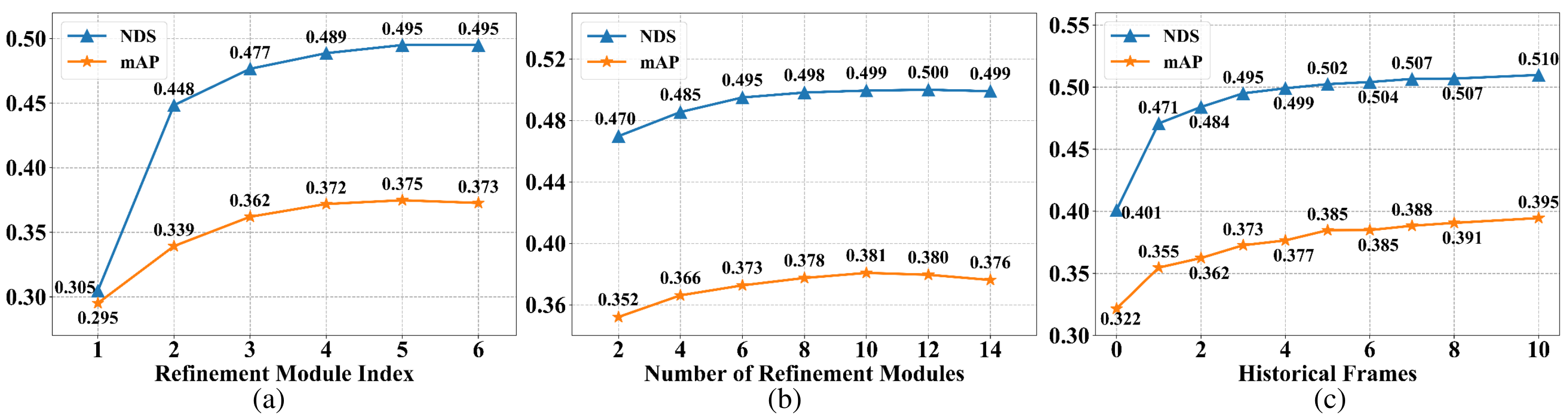}
  \caption{Ablation study of the influence of Refinement Modules and Historical Frame. In this experiment, the input image size is set to $320\times 800$ and learnable keypoints are removed.}
  \label{fig:ablationdeocer}
\end{figure*}

%-------------------------------------------------------------------------
\subsection{Datasets and Metrics}
    We evaluate our method on the nuScenes benchmark. The nuScenes dataset~\cite{caesar2020nuscenes} contains data for 1000 scenes, of which 700, 150, and 150 scenes are used for training, validation, and testing, respectively. Each scene is a 20 second video clip at 2 frames per second. Each frame has image data from 6 cameras, and enough annotations such as the category, 3D bounding box, and ID of objects.

For the 3D detection task, evaluation metrics include mean Average Precision (mAP), mean Average Error of Translation (mATE), Scale (mASE),  Orientation  (mAOE),  Velocity (mAVE),  Attribute (mAAE) and  nuScenes Detection Score (NDS), where NDS is a weighted average of other metrics.
    For the object tracking task, Average Multi-Object Tracking Accuracy (AMOTA), Average Multi-Object Tracking Precision (AMOTP) and Recall are the three main evaluation metrics. Please refer to ~\cite{caesar2020nuscenes,weng2019baseline} for details.
%-------------------------------------------------------------------------
\begin{table}
  \centering
  \begin{tabular}{@{}c|cc|cc@{}}
    \toprule
    Method & mAP$\uparrow$  & NDS$\uparrow$ & \footnotesize{FLOPs(G)} & \footnotesize{Params(M)} \\
    \midrule
    DETR3D & 0.346  & 0.425 & 996.6 & 53.3 \\
    LS-DETR* & 0.348 & 0.397 & 1087.7 & 74.0 \\
    BEVFormer-S & 0.375  & 0.448 & 1303.5 & 68.7 \\
    Sparse4D$_{T=1}$ & 0.382  & 0.451 & 1019.2 & 58.1  \\
    Sparse4D$_{T=4}$ & \textbf{0.436} & \textbf{0.541} & 1113.8 & 58.9 \\
    \bottomrule
  \end{tabular}
  \caption{Comparison of FLOPs and parameter amount of different methods, where all methods share same backbone with 854.3G Flops and 23.3M parameters. * LS-DETR here stands for Lift-Splat + Deformable DETR.}
  \label{tab:flops-params}
\end{table}

\begin{table*}
  \centering
  \begin{tabular}{@{}l|cc|cccccc|c@{}}
    \toprule
    Method & Temporal & Backbone & mAP$\uparrow$ & mATE$\downarrow$ & mASE$\downarrow$ & mAOE$\downarrow$ & mAVE$\downarrow$ & mAAE$\downarrow$ & NDS$\uparrow$ \\
    \midrule
    FCOS3D & \XSolidBrush & ResNet101 & 0.299 & 0.785 & \textbf{0.268} & 0.557 & 1.396 & \textbf{0.154} & 0.373 \\
    DETR3D & \XSolidBrush & ResNet101 & 0.349 & 0.716 & \textbf{0.268} & \textbf{0.379} & 0.842 & 0.200 & 0.434  \\
    BEVDet & \XSolidBrush & ResNet101 & 0.357 & \textbf{0.710} & 0.270 & 0.490 & 0.885 & 0.224 & 0.421 \\
    BEVFormer-S & \XSolidBrush & ResNet101 & 0.375 & 0.725 & 0.272 & 0.391 & \textbf{0.802} & 0.200 & 0.448 \\
    \midrule
    Sparse4D$_{T=1}$ & \XSolidBrush & ResNet101 & \textbf{0.382} & \textbf{0.710} & 0.279 & 0.411 & 0.806 & 0.196 & \textbf{0.451}\\
    \midrule
    BEVFormer & \checkmark & ResNet101 & 0.416 & 0.673 & 0.274 & 0.372 & 0.394 & 0.198 & 0.517 \\
    BEVDet4D & \checkmark & Swin-Base & 0.396 & 0.619 & \textbf{0.260} & 0.361 & 0.399 & 0.189 & 0.515 \\
    PolarFormer-T & \checkmark & ResNet101 & 0.432 & 0.648 & 0.270 & \textbf{0.348} & 0.409 & 0.201 & 0.528\\
    BEVDepth & \checkmark & ResNet101 & 0.412 & \textbf{0.565} & 0.266 & 0.358 & 0.331 & 0.190 & 0.535 \\
    \midrule
    Sparse4D$_{T=4}$ & \checkmark & ResNet101 & 0.436 & 0.633 & 0.279 & 0.363 & 0.317 & \textbf{0.177} & 0.541 \\
    Sparse4D$_{T=9(-6)}$ & \checkmark & ResNet101 & \textbf{0.445} & 0.613 & 0.279 & 0.378 & \textbf{0.303} & 0.180 & 0.547 \\
    Sparse4D$_{T=9(-6)}^{\dagger}$ & \checkmark & ResNet101 & 0.444 & 0.603 & 0.276 & 0.360 & 0.309 & 0.178 & \textbf{0.550} \\
    \bottomrule
  \end{tabular}
  \caption{Results of 3D object detection on nuScenes validation dataset.$\dagger$ indicates the number of training epochs is 48.  The subscript $T=X(-Y)$ means that the number of frames used is $X$, and the feature $f_{t}'$ of $Y$ historical frames is randomly detached during training. }
  \label{tab:detectionval}
\end{table*}

\subsection{Implementation Details}

The initial $\left\{x, y, z\right\}$ parameters of the 3D anchors are obtained by performing K-Means clustering on the training set, and the other parameters are all initialized with fixed values $\left\{1, 1, 1, 0, 1, 0, 0, 0\right\}$. The instance feature uses random initialization.  By default, the number of 3D anchors and instance features $M$ is set to 900, the number of cascade refinement modules is 6, the number of feature map scales $S$ from the neck is 4, the number of fixed keypoints $K_F$ is 7, the number of learnable keypoints $K_L$ is 6, the input image size is $640 \times 1600$, and the backbone is ResNet101.

Sparse4D is trained with AdamW optimizer~\cite{adamw}. The initial learning rates of backbone and other parameters are 2e-5 and 2e-4, respectively. The decay strategy is cosine annealing~\cite{loshchilov2016sgdr}. The initial network parameters come from pre-trained FCOS3D~\cite{wang2021fcos3d}. For the experiments on the nuScenes test set, the network was trained for 48 epochs, and the rest of the experiments were only trained for 24 epochs unless otherwise specified.
%$\left\{F_{t}|t_s\le t < t_0\right\}$
In order to save GPU memory, we detach the feature maps  of all historical frames and the fusion features $f_{t}'$ of a random part of historical frames during the training phase. CBGS~\cite{zhu2019cbgs} and test time augmentation were not used in all experiments.
%-------------------------------------------------------------------------
\subsection{Ablation Studies and Analysis}

\begin{table*}
  \centering
  \begin{tabular}{@{}l|cc|cccccc|c@{}}
    \toprule
    Method & Sparse & Backbone & mAP$\uparrow$ & mATE$\downarrow$ & mASE$\downarrow$ & mAOE$\downarrow$ & mAVE$\downarrow$ & mAAE$\downarrow$ & NDS$\uparrow$ \\
    \midrule
    SRCN3D & \checkmark &  VoVNet-99 & 0.396 & 0.673 & 0.269 & 0.403 & 0.875 & 0.129 & 0.463\\
    DETR3D & \checkmark & VoVNet-99 & 0.412 & 0.641 & 0.255 & 0.394 & 0.845 & 0.133 & 0.479\\
    Graph-DETR3D & \checkmark & VoVNet-99 & 0.425 & 0.621 & 0.251 & 0.386 & 0.790 & 0.128 & 0.495 \\
    \midrule
    UVTR  & \XSolidBrush & VoVNet-99 & 0.472 & 0.577 & 0.253 & 0.391 & 0.508 & 0.123 & 0.551 \\
    BEVDet4D & \XSolidBrush & Swin-Base & 0.451 & 0.511 & \textbf{0.241} & 0.386 & \textbf{0.301} & 0.121 & 0.569 \\
    BEVFormer & \XSolidBrush & VoVNet-99 & 0.481 & 0.582 & 0.256 & 0.375 & 0.378 & 0.126 & 0.569 \\
    PETRv2 & \XSolidBrush & VoVNet-99 & 0.490 & 0.561 & 0.243 & \textbf{0.361} & 0.343 & \textbf{0.120} & 0.582 \\
    BEVDistill & \XSolidBrush & ConvNeXt-Base & 0.496 & \textbf{0.475} & 0.249 & 0.378 & 0.313 & 0.125 & 0.594\\ \midrule
    Sparse4D$_{T=9(-6)}^{\dagger}$ & \checkmark & VoVNet-99 & \textbf{0.511} & 0.533 & 0.263 & 0.369 & 0.317 & 0.124 & \textbf{0.595}\\
    \bottomrule
  \end{tabular}
  \caption{Results of 3D object detection on nuScenes test dataset. The superscript $\dagger$ and subscript $T=X(-Y)$ have the same meaning as in \cref{tab:detectionval}. Here the initialization parameters of VoVNet-99 are all pre-trained from DD3D~\cite{DD3D} with extra data.}
  \label{tab:detectiontest}
\end{table*}

\textbf{Depth Reweight Module and Learnable Keypoints.} By adding the depth reweight module or learnable keypoints, we compare and analyze the before-and-after changes in metrics on the nuScenes validation dataset, see \cref{tab:ablation-DRM-LKP}. It can be seen that the addition of these two modules has a certain promotion effect on the model performance, and the impact on the metric NDS is similar, $0.33\%$ and $0.35\%$, respectively. When these two structures are added together, all metrics will increase, among which mAP increases by $0.38\%$ and NDS increases by $0.79\%$.

\textbf{Motion Compensation.} When generating 4D keypoints, we consider both the ego vehicle motion and the object motion. From \cref{tab:ablationmotion}, we can see that even without any motion compensation, after adding temporal information, the model performance still improved to a certain extent, in which mAVE increased by $20.8\%$ and NDS increased by $2.3\%$.
However, the overall perceptual performance of this model is still low. After adding ego motion compensation, the model effect is significantly improved, especially mAP and mAVE, which are increased by $4.2\%$ and $28.4\%$ respectively, and the comprehensive metric NDS is increased by $6.4\%$. On this basis, considering the motion of the object to be detected, the detection accuracy is not improved, but the error of the speed estimation will be reduced by $6.9\%$, thus increasing the NDS by about $0.7\%$.

\textbf{Number of Refinements.} The number of iterative refinements also has a significant impact on detection performance. In this regard, we designed two sets of experiments for analysis. In the first set of experiments, we train a model with 6 refinement modules and compute the detection metrics output by each refinement module. As can be seen from \cref{fig:ablationdeocer}(a), as the times of refinements increases, the overall metrics show an increasing trend, and the growth rate gradually decreases. Compared with the first one, the output accuracy of the second refinement module is significantly increased, but the detection effect between the fifth refinement module and the sixth module is not much different. In the second set of experiments, we train multiple models whose number of refinement modules is increased from 2 to 14. When the number of modules is 10, the NDS is the highest at $38.1\%$, as shown in \cref{fig:ablationdeocer}(b).

\textbf{Number of Historical Frames.} We train and infer Sparse4D with varying numbers of historical frames and find that model performance continues to grow as the number of frames increases(\cref{fig:ablationdeocer}(c)). Even if the number of frames increases to 10 (equivalent to 5 seconds in history), there is still a small increase compared to 8 frames. There may still be room for improvement in Sparse4D's performance if more frames are added. However, due to the limitations of our training device's memory (V100, 32G), it was not possible to try more frames.

\textbf{FLOPs and Parameters.}
In this experiment, the input image size is set to 900x1600 and  ResNet101 is used as backbone, and the experimental results are shown in \cref{tab:flops-params}.
When $T=1$, the FLOPs of our model is 1019.2G, and the parameter amount is 58.1M. Compared with DETR3D, the amount of calculation  and parameter are only increased by $2.3\%$ and $9.0\%$, respectively, and the mAP and NDS are increased by $3.6\%$ and $2.6\%$.  Compared with Lift-splat and BEVFormer-S, we have certain advantages in algorithm metrics, FLOPs and parameter amount. After adding 3 history frames, Sparse4D with temporal fusion only increases $9.3\%$ FLOPs and $1.4\%$ parameters, and achieves a very noticeable improvement, $5.4\%$ mAP and $9.0\%$ NDS.

%-------------------------------------------------------------------------
\subsection{Main Results}
%\textbf{Validation.}
The comparison of the results on the nuScenes validation set is shown in \cref{tab:detectionval}. Among all the non-temporal models, Sparse4D gets the highest mAP and NDS. Compared with the baseline of the sparse methods, DERT3D, we improve mAP and NDS by $3.3\%$ and $1.7\%$, respectively. Compared to the baseline of the BEV-based methods, BEVFormer, we lead by $0.7\%$ on mAP and $0.3\%$ on NDS.
We also compared Sparse4D with other SOTA temporal algorithms, and still obtained the best NDS and mAP. When $T=4$, Sparse4D outperforms BEVDepth by $2.4\%$ on mAP and $0.6\%$ on NDS. When $T$ is increased from 4 to 9, the mAP and NDS of Sparse4D are improved by $0.9\%$ and $0.6\%$ respectively. Moreover, adding 24 epochs to training can further improve NDS by $0.3\%$.

%\textbf{Test.}
We compare Sparse4D with other SOTA algorithms on the nuScenes test set (online leaderboard). As shown in the \cref{tab:detectiontest}, with DD3D~\cite{DD3D} pre-trained VoVNet-99, Sparse4D achieves $51.1\%$ and $59.5\%$ on mAP and NDS metrics, respectively, outperforming all non-BEV methods including PETRv2. Compared with baseline DETR3D, our method has achieved significant improvements. The mAP and NDS have increased by $9.9\%$ and $11.6\%$, respectively, greatly improving the competitiveness of sparse methods. In addition, Sparse4D is also superior to the dense BEV based methods including UVTR~\cite{li2022unifying}, BEVdet~\cite{huang2022bevdet4d}, BEVFormer~\cite{bevformer} and BEVDistill~\cite{anonymous2023bevdistill}, especially in mAP, which is $1.5\%$ higher than BEVDistill.
%-------------------------------------------------------------------------
\subsection{Extend to 3D Object Tracking}
    Based on the tracking-by-detection framework~\cite{trackingsurvey}, Sparse4D is easily extended to a tracker. We use the instance features and bounding boxes output by the last refinement module to extract identity features, and use a lightweight sub-network to estimate the correlation matrix between historical trajectories and current objects. Then, the matching relationship between the historical trajectory and the current object will be obtained using the Hungarian matching algorithm. As shown in \cref{tab:trackingtest}, Sparse4D obtains 0.519 AMOTA and 1.078 AMOTP on nuScenes test set, which is ahead of most learning-based methods.

\begin{table}
  \centering
  \begin{tabular}{@{}l|cccccccc@{}}
    \toprule
    Method & AMOTA$\uparrow$ & AMOTP$\downarrow$ & Recall$\uparrow$ \\
    \midrule
    MUTR3D~\cite{zhang2022mutr3d} & 0.270 & 1.494 & 0.411\\
    PolarDETR~\cite{polardetr} & 0.273 & 1.185 & 0.404\\
    % BEVTrack & 0.341 & 1.107 & 0.463\\
    SRCN3D & 0.398 & 1.317 & 0.538\\
    % XTracker & 0.430 & 1.196 & 0.525 \\
    DAMEN-T & 0.460 & 1.155 & 0.558\\
    QTrack~\cite{qtrack} & 0.480 & 1.100 & 0.583 \\
    UVTR-GreedyTrack & \textbf{0.519} & 1.125 & 0.599 \\
    \midrule
    Sparse4D & \textbf{0.519} & \textbf{1.078} & \textbf{0.633} \\
    \bottomrule
  \end{tabular}
  \caption{Results of 3D multi-object tracking on nuScenes test set.}
  \label{tab:trackingtest}
\end{table}
%------------------------------------------------------------------------
\section{Conclusion}
In this work, we propose a new method, Sparse4D, which achieves feature-level fusion of multi-timestamp and multi-view through a deformable 4D aggregation module, and uses iterative refinement to achieve 3D box regression. Sparse4D can provide excellent perceptual performance, and it outperforms all existing sparse algorithms and most BEV-based algorithms on the nuScenes leaderboard.

We believe that Sparse4D still has a lot of room for improvement. For example, in the depth reweight module, multi-view stereo (MVS)~\cite{yao2018mvsnet,li2022bevstereo} technology can be added to obtain more accurate depth. Camera parameters can also be considered in the encoder to improve 3D generalization~\cite{camconv,li2022bevdepth}. Therefore, we hope that Sparse4D can become a new baseline for sparse 3D detection. In addition, the framework of Sparse4D can also be extended to other tasks, such as HD map construction, occupancy estimation, 3D reconstruction, etc.
%%%%%%%%% REFERENCES
{\small
\bibliographystyle{ieee_fullname}
\bibliography{egbib.bib}

\begin{thebibliography}{10}\itemsep=-1pt

\bibitem{anonymous2023bevdistill}
Anonymous.
\newblock {BEVD}istill: Cross-modal {BEV} distillation for multi-view 3d object
  detection.
\newblock In {\em Submitted to The Eleventh International Conference on
  Learning Representations}, 2023.
\newblock under review.

\bibitem{caesar2020nuscenes}
Holger Caesar, Varun Bankiti, Alex~H Lang, Sourabh Vora, Venice~Erin Liong,
  Qiang Xu, Anush Krishnan, Yu Pan, Giancarlo Baldan, and Oscar Beijbom.
\newblock nuscenes: A multimodal dataset for autonomous driving.
\newblock In {\em Proceedings of the IEEE/CVF conference on computer vision and
  pattern recognition}, pages 11621--11631, 2020.

\bibitem{DETR}
Nicolas Carion, Francisco Massa, Gabriel Synnaeve, Nicolas Usunier, Alexander
  Kirillov, and Sergey Zagoruyko.
\newblock End-to-end object detection with transformers.
\newblock In {\em European conference on computer vision}, pages 213--229.
  Springer, 2020.

\bibitem{polardetr}
Shaoyu Chen, Xinggang Wang, Tianheng Cheng, Qian Zhang, Chang Huang, and Wenyu
  Liu.
\newblock Polar parametrization for vision-based surround-view 3d detection.
\newblock {\em arXiv preprint arXiv:2206.10965}, 2022.

\bibitem{chen2022graph}
Zehui Chen, Zhenyu Li, Shiquan Zhang, Liangji Fang, Qinhong Jiang, and Feng
  Zhao.
\newblock Graph-detr3d: Rethinking overlapping regions for multi-view 3d object
  detection.
\newblock {\em arXiv preprint arXiv:2204.11582}, 2022.

\bibitem{trackingsurvey}
Gioele Ciaparrone, Francisco~Luque S{\'a}nchez, Siham Tabik, Luigi Troiano,
  Roberto Tagliaferri, and Francisco Herrera.
\newblock Deep learning in video multi-object tracking: A survey.
\newblock {\em Neurocomputing}, 381:61--88, 2020.

\bibitem{duan2019centernet}
Kaiwen Duan, Song Bai, Lingxi Xie, Honggang Qi, Qingming Huang, and Qi Tian.
\newblock Centernet: Keypoint triplets for object detection.
\newblock In {\em Proceedings of the IEEE/CVF international conference on
  computer vision}, pages 6569--6578, 2019.

\bibitem{camconv}
Jose~M Facil, Benjamin Ummenhofer, Huizhong Zhou, Luis Montesano, Thomas Brox,
  and Javier Civera.
\newblock Cam-convs: Camera-aware multi-scale convolutions for single-view
  depth.
\newblock In {\em Proceedings of the IEEE/CVF Conference on Computer Vision and
  Pattern Recognition}, pages 11826--11835, 2019.

\bibitem{he2017mask}
Kaiming He, Georgia Gkioxari, Piotr Doll{\'a}r, and Ross Girshick.
\newblock Mask r-cnn.
\newblock In {\em Proceedings of the IEEE international conference on computer
  vision}, pages 2961--2969, 2017.

\bibitem{resnet}
Kaiming He, Xiangyu Zhang, Shaoqing Ren, and Jian Sun.
\newblock Deep residual learning for image recognition.
\newblock In {\em Proceedings of the IEEE conference on computer vision and
  pattern recognition}, pages 770--778, 2016.

\bibitem{huang2022bevdet4d}
Junjie Huang and Guan Huang.
\newblock Bevdet4d: Exploit temporal cues in multi-camera 3d object detection.
\newblock {\em arXiv preprint arXiv:2203.17054}, 2022.

\bibitem{huang2021bevdet}
Junjie Huang, Guan Huang, Zheng Zhu, and Dalong Du.
\newblock Bevdet: High-performance multi-camera 3d object detection in
  bird-eye-view.
\newblock {\em arXiv preprint arXiv:2112.11790}, 2021.

\bibitem{krizhevsky2017imagenet}
Alex Krizhevsky, Ilya Sutskever, and Geoffrey~E Hinton.
\newblock Imagenet classification with deep convolutional neural networks.
\newblock {\em Communications of the ACM}, 60(6):84--90, 2017.

\bibitem{vovnet}
Youngwan Lee, Joong-won Hwang, Sangrok Lee, Yuseok Bae, and Jongyoul Park.
\newblock An energy and gpu-computation efficient backbone network for
  real-time object detection.
\newblock In {\em Proceedings of the IEEE/CVF conference on computer vision and
  pattern recognition workshops}, pages 0--0, 2019.

\bibitem{li2022bevstereo}
Yinhao Li, Han Bao, Zheng Ge, Jinrong Yang, Jianjian Sun, and Zeming Li.
\newblock Bevstereo: Enhancing depth estimation in multi-view 3d object
  detection with dynamic temporal stereo.
\newblock {\em arXiv preprint arXiv:2209.10248}, 2022.

\bibitem{li2022unifying}
Yanwei Li, Yilun Chen, Xiaojuan Qi, Zeming Li, Jian Sun, and Jiaya Jia.
\newblock Unifying voxel-based representation with transformer for 3d object
  detection.
\newblock {\em arXiv preprint arXiv:2206.00630}, 2022.

\bibitem{li2022bevdepth}
Yinhao Li, Zheng Ge, Guanyi Yu, Jinrong Yang, Zengran Wang, Yukang Shi,
  Jianjian Sun, and Zeming Li.
\newblock Bevdepth: Acquisition of reliable depth for multi-view 3d object
  detection.
\newblock {\em arXiv preprint arXiv:2206.10092}, 2022.

\bibitem{bevformer}
Zhiqi Li, Wenhai Wang, Hongyang Li, Enze Xie, Chonghao Sima, Tong Lu, Qiao Yu,
  and Jifeng Dai.
\newblock Bevformer: Learning bird's-eye-view representation from multi-camera
  images via spatiotemporal transformers.
\newblock {\em arXiv preprint arXiv:2203.17270}, 2022.

\bibitem{liang2022bevfusion}
Tingting Liang, Hongwei Xie, Kaicheng Yu, Zhongyu Xia, Zhiwei Lin, Yongtao
  Wang, Tao Tang, Bing Wang, and Zhi Tang.
\newblock Bevfusion: A simple and robust lidar-camera fusion framework.
\newblock {\em arXiv preprint arXiv:2205.13790}, 2022.

\bibitem{fpn}
Tsung-Yi Lin, Piotr Doll{\'a}r, Ross Girshick, Kaiming He, Bharath Hariharan,
  and Serge Belongie.
\newblock Feature pyramid networks for object detection.
\newblock In {\em Proceedings of the IEEE conference on computer vision and
  pattern recognition}, pages 2117--2125, 2017.

\bibitem{lin2017focal}
Tsung-Yi Lin, Priya Goyal, Ross Girshick, Kaiming He, and Piotr Doll{\'a}r.
\newblock Focal loss for dense object detection.
\newblock In {\em Proceedings of the IEEE international conference on computer
  vision}, pages 2980--2988, 2017.

\bibitem{liu2016ssd}
Wei Liu, Dragomir Anguelov, Dumitru Erhan, Christian Szegedy, Scott Reed,
  Cheng-Yang Fu, and Alexander~C Berg.
\newblock Ssd: Single shot multibox detector.
\newblock In {\em European conference on computer vision}, pages 21--37.
  Springer, 2016.

\bibitem{liu2022petr}
Yingfei Liu, Tiancai Wang, Xiangyu Zhang, and Jian Sun.
\newblock Petr: Position embedding transformation for multi-view 3d object
  detection.
\newblock {\em arXiv preprint arXiv:2203.05625}, 2022.

\bibitem{liu2022petrv2}
Yingfei Liu, Junjie Yan, Fan Jia, Shuailin Li, Qi Gao, Tiancai Wang, Xiangyu
  Zhang, and Jian Sun.
\newblock Petrv2: A unified framework for 3d perception from multi-camera
  images.
\newblock {\em arXiv preprint arXiv:2206.01256}, 2022.

\bibitem{liu2020smoke}
Zechen Liu, Zizhang Wu, and Roland T{\'o}th.
\newblock Smoke: Single-stage monocular 3d object detection via keypoint
  estimation.
\newblock In {\em Proceedings of the IEEE/CVF Conference on Computer Vision and
  Pattern Recognition Workshops}, pages 996--997, 2020.

\bibitem{loshchilov2016sgdr}
Ilya Loshchilov and Frank Hutter.
\newblock Sgdr: Stochastic gradient descent with warm restarts.
\newblock {\em arXiv preprint arXiv:1608.03983}, 2016.

\bibitem{adamw}
Ilya Loshchilov and Frank Hutter.
\newblock Decoupled weight decay regularization.
\newblock {\em arXiv preprint arXiv:1711.05101}, 2017.

\bibitem{DD3D}
Dennis Park, Rares Ambrus, Vitor Guizilini, Jie Li, and Adrien Gaidon.
\newblock Is pseudo-lidar needed for monocular 3d object detection?
\newblock In {\em Proceedings of the IEEE/CVF International Conference on
  Computer Vision}, pages 3142--3152, 2021.

\bibitem{solofusion}
Jinhyung Park, Chenfeng Xu, Shijia Yang, Kurt Keutzer, Kris Kitani, Masayoshi
  Tomizuka, and Wei Zhan.
\newblock Time will tell: New outlooks and a baseline for temporal multi-view
  3d object detection.
\newblock {\em arXiv preprint arXiv:2210.02443}, 2022.

\bibitem{philion2020lift}
Jonah Philion and Sanja Fidler.
\newblock Lift, splat, shoot: Encoding images from arbitrary camera rigs by
  implicitly unprojecting to 3d.
\newblock In {\em European Conference on Computer Vision}, pages 194--210.
  Springer, 2020.

\bibitem{qian2020end2endpseudo}
Rui Qian, Divyansh Garg, Yan Wang, Yurong You, Serge Belongie, Bharath
  Hariharan, Mark Campbell, Kilian~Q Weinberger, and Wei-Lun Chao.
\newblock End-to-end pseudo-lidar for image-based 3d object detection.
\newblock In {\em Proceedings of the IEEE/CVF Conference on Computer Vision and
  Pattern Recognition}, pages 5881--5890, 2020.

\bibitem{reading2021categorical}
Cody Reading, Ali Harakeh, Julia Chae, and Steven~L Waslander.
\newblock Categorical depth distribution network for monocular 3d object
  detection.
\newblock In {\em Proceedings of the IEEE/CVF Conference on Computer Vision and
  Pattern Recognition}, pages 8555--8564, 2021.

\bibitem{ren2015faster}
Shaoqing Ren, Kaiming He, Ross Girshick, and Jian Sun.
\newblock Faster r-cnn: Towards real-time object detection with region proposal
  networks.
\newblock {\em Advances in neural information processing systems}, 28, 2015.

\bibitem{OFT}
Thomas Roddick, Alex Kendall, and Roberto Cipolla.
\newblock Orthographic feature transform for monocular 3d object detection.
\newblock {\em arXiv preprint arXiv:1811.08188}, 2018.

\bibitem{shi2022srcn3d}
Yining Shi, Jingyan Shen, Yifan Sun, Yunlong Wang, Jiaxin Li, Shiqi Sun, Kun
  Jiang, and Diange Yang.
\newblock Srcn3d: Sparse r-cnn 3d surround-view camera object detection and
  tracking for autonomous driving.
\newblock {\em arXiv preprint arXiv:2206.14451}, 2022.

\bibitem{sparsercnn}
Peize Sun, Rufeng Zhang, Yi Jiang, Tao Kong, Chenfeng Xu, Wei Zhan, Masayoshi
  Tomizuka, Lei Li, Zehuan Yuan, Changhu Wang, et~al.
\newblock Sparse r-cnn: End-to-end object detection with learnable proposals.
\newblock In {\em Proceedings of the IEEE/CVF conference on computer vision and
  pattern recognition}, pages 14454--14463, 2021.

\bibitem{tan2020efficientdet}
Mingxing Tan, Ruoming Pang, and Quoc~V Le.
\newblock Efficientdet: Scalable and efficient object detection.
\newblock In {\em Proceedings of the IEEE/CVF conference on computer vision and
  pattern recognition}, pages 10781--10790, 2020.

\bibitem{tian2019fcos}
Zhi Tian, Chunhua Shen, Hao Chen, and Tong He.
\newblock Fcos: Fully convolutional one-stage object detection.
\newblock In {\em Proceedings of the IEEE/CVF international conference on
  computer vision}, pages 9627--9636, 2019.

\bibitem{wang2021fcos3d}
Tai Wang, Xinge Zhu, Jiangmiao Pang, and Dahua Lin.
\newblock Fcos3d: Fully convolutional one-stage monocular 3d object detection.
\newblock In {\em Proceedings of the IEEE/CVF International Conference on
  Computer Vision}, pages 913--922, 2021.

\bibitem{wang2019pseudo}
Yan Wang, Wei-Lun Chao, Divyansh Garg, Bharath Hariharan, Mark Campbell, and
  Kilian~Q Weinberger.
\newblock Pseudo-lidar from visual depth estimation: Bridging the gap in 3d
  object detection for autonomous driving.
\newblock In {\em Proceedings of the IEEE/CVF Conference on Computer Vision and
  Pattern Recognition}, pages 8445--8453, 2019.

\bibitem{wang2022detr3d}
Yue Wang, Vitor~Campagnolo Guizilini, Tianyuan Zhang, Yilun Wang, Hang Zhao,
  and Justin Solomon.
\newblock Detr3d: 3d object detection from multi-view images via 3d-to-2d
  queries.
\newblock In {\em Conference on Robot Learning}, pages 180--191. PMLR, 2022.

\bibitem{weng2019baseline}
Xinshuo Weng and Kris Kitani.
\newblock A baseline for 3d multi-object tracking.
\newblock {\em arXiv preprint arXiv:1907.03961}, 1(2):6, 2019.

\bibitem{weng2019monocularpseudo}
Xinshuo Weng and Kris Kitani.
\newblock Monocular 3d object detection with pseudo-lidar point cloud.
\newblock In {\em Proceedings of the IEEE/CVF International Conference on
  Computer Vision Workshops}, pages 0--0, 2019.

\bibitem{qtrack}
Jinrong Yang, En Yu, Zeming Li, Xiaoping Li, and Wenbing Tao.
\newblock Quality matters: Embracing quality clues for robust 3d multi-object
  tracking.
\newblock {\em arXiv preprint arXiv:2208.10976}, 2022.

\bibitem{yao2018mvsnet}
Yao Yao, Zixin Luo, Shiwei Li, Tian Fang, and Long Quan.
\newblock Mvsnet: Depth inference for unstructured multi-view stereo.
\newblock In {\em Proceedings of the European conference on computer vision
  (ECCV)}, pages 767--783, 2018.

\bibitem{zhang2022monodetr}
Renrui Zhang, Han Qiu, Tai Wang, Xuanzhuo Xu, Ziyu Guo, Yu Qiao, Peng Gao, and
  Hongsheng Li.
\newblock Monodetr: Depth-aware transformer for monocular 3d object detection.
\newblock {\em arXiv preprint arXiv:2203.13310}, 2022.

\bibitem{zhang2022mutr3d}
Tianyuan Zhang, Xuanyao Chen, Yue Wang, Yilun Wang, and Hang Zhao.
\newblock Mutr3d: A multi-camera tracking framework via 3d-to-2d queries.
\newblock In {\em Proceedings of the IEEE/CVF Conference on Computer Vision and
  Pattern Recognition}, pages 4537--4546, 2022.

\bibitem{zhang2022simple}
Yunpeng Zhang, Wenzhao Zheng, Zheng Zhu, Guan Huang, Jie Zhou, and Jiwen Lu.
\newblock A simple baseline for multi-camera 3d object detection.
\newblock {\em arXiv preprint arXiv:2208.10035}, 2022.

\bibitem{zhang2022beverse}
Yunpeng Zhang, Zheng Zhu, Wenzhao Zheng, Junjie Huang, Guan Huang, Jie Zhou,
  and Jiwen Lu.
\newblock Beverse: Unified perception and prediction in birds-eye-view for
  vision-centric autonomous driving.
\newblock {\em arXiv preprint arXiv:2205.09743}, 2022.

\bibitem{zhu2019cbgs}
Benjin Zhu, Zhengkai Jiang, Xiangxin Zhou, Zeming Li, and Gang Yu.
\newblock Class-balanced grouping and sampling for point cloud 3d object
  detection.
\newblock {\em arXiv preprint arXiv:1908.09492}, 2019.

\bibitem{deformableDETR}
Xizhou Zhu, Weijie Su, Lewei Lu, Bin Li, Xiaogang Wang, and Jifeng Dai.
\newblock Deformable detr: Deformable transformers for end-to-end object
  detection.
\newblock {\em arXiv preprint arXiv:2010.04159}, 2020.

\end{thebibliography}
}

\end{document}